\documentclass{article}

\usepackage{arxiv}

\usepackage[utf8]{inputenc} % allow utf-8 input
\usepackage[T1]{fontenc}    % use 8-bit T1 fonts
\usepackage{hyperref}       % hyperlinks
\usepackage{url}            % simple URL typesetting
\usepackage{booktabs}       % professional-quality tables
\usepackage{amsfonts}       % blackboard math symbols
\usepackage{nicefrac}       % compact symbols for 1/2, etc.
\usepackage{microtype}      % microtypography
\usepackage{lipsum}
\usepackage{graphicx}
\usepackage{color,soul}
\usepackage{subcaption}
\usepackage{xcolor}
\usepackage[noadjust]{cite}
\graphicspath{ {./images/} }

\title{Automatically identifying a mobile phone user's position within a vehicle}

\author{
 Matt Knutson \\
  Minnesota HealthSolutions\\
  Minneapolis, MN 55413 \\
  \texttt{matt@minnhealth.com} \\
  
   \And

 Kevin Kramer \\
  Minnesota HealthSolutions\\
  Minneapolis, MN 55413 \\
  \texttt{kevin@minnhealth.com} \\
  
   \And

 Sara Seifert \\
  Minnesota HealthSolutions\\
  Minneapolis, MN 55413 \\
  \texttt{sara@minnhealth.com} \\
  
   \And

 Ryan Chamberlain \\
  Minnesota HealthSolutions\\
  Minneapolis, MN 55413 \\
  \texttt{ryan@minnhealth.com} \\
}

\begin{document}
\maketitle
\begin{abstract}
 Traffic-related injuries and fatalities are major health risks in the United States. Mobile phone use while driving quadruples the risk for a motor vehicle crash. This work demonstrates the feasibility of using the mobile phone camera to passively detect the location of the phone's user within a vehicle. In a large, varied dataset we were able correctly identify if the user was in the driver's seat or one of the passenger seats with 94.9\% accuracy. This model could be used by application developers to selectively change or lock functionality while a user is driving, but not if the user is a passenger in a moving vehicle.
\end{abstract}

% keywords can be removed
%\keywords{First keyword \and Second keyword \and More}

\section{Introduction}
Traffic-related injuries and fatalities are major health risks in the United States. Mobile phone use while driving (MPUWD) quadruples the risk for a motor vehicle crash \cite{Redelmeier1997-fj}. In 2019 alone, 3,142 people died in crashes with distracted drivers \cite{nhtsa-dd}. Drivers using a smartphone have reduced ability to detect and respond to roadway stimuli and events \cite{Caird2008-tu, Fitch2013-vh, Handel2014-ku, Klauer2014-tf, Strayer2013-tp}. Teens in general have a stronger tendency to use their mobile phones \cite{Walsh2008-rd}, and studies have shown that nearly half of adolescent drivers engage in MPUWD \cite{Curry2011-cf, Neyens2008-sn}. Legal bans on phone use while driving have shown only modest results, likely due to the difficulties of enforcement \cite{Delgado2016-sr}. Modern mobile phones have built-in functionality to detect when a phone is in a moving vehicle; however, those libraries are unable to differentiate between drivers and passengers, and commercially available MPUWD software typically includes a simple option for users to manually disable it during drives where they are a passenger.

\paragraph{Related work.} To overcome this limitation on current phones researchers have utilized vehicle additions or modifications to differentiate between drivers and passengers. Previous efforts to determine the seating location of a phone user have relied on either Bluetooth signals \cite{Yuan2020-np, Yang2011-sh}, phone sensor data \cite{Park2018-py}, or car sensor data \cite{Kar2017-ew}. Many of these methods rely on either installing Bluetooth beacons, having the user's phone paired to the car's Bluetooth radio, or access the car's sensor bus, and the accuracy has not exceeded 93\%. Another method is to use the phone's camera to determine the user's location using computer vision \cite{Khurana2020-qc}. This work was able to reach 94\% accuracy, but in a a limited dataset. Camera-based methods require no hardware installation or Bluetooth pairing by using only the user's phone camera to determine the user location within the car. We are expanding on this work by demonstrating the feasibility of a computer vision method in a larger, more varied, dataset. 

\section{Methodology}

\paragraph{Dataset.} The seat detection dataset was collected by at a number of used car dealerships. A custom mobile phone app was written to assist users in taking a series of pictures from each seat of the vehicle with prompts for phone orientation. Multiple images were acquired from each seat using slightly different orientations. Images were acquired from driver seats as well as front and rear passenger seats. The make, model, and year of each car was recorded along with the time of day. The image acquisition times varied between full sun, overcast, dawn, dusk and night. A total of 12,042 images were acquired from 120 cars and trucks. There were 3,721 images from the driver seat and 8,321 from the passenger seats. The model years ranged from 1996 - 2020. The dataset was randomly split into training, development, and testing sets using the car as the split key, so the same car did not appear in multiple sets.

\paragraph{CNN architecture.} The neural network was implemented using Keras and Tensorflow. The architecture was MobileNet-V2 \cite{Sandler2018-ix} feature extractor provided with Keras pretrained using the ImageNet dataset \cite{Deng2009-ov}. A custom head was attached to the MobileNet feature extractor that consisted of: 1) a $1 \times 1$ convolution followed by batch normalization \cite{Ioffe2015-il}, a ReLU activation, and dropout, 2) a $7 \times 7$ convolution followed by batch normalization, a ReLU activation, and dropout, and 3) a global max-pooling and a final fully connected layer.

\paragraph{Training procedure.} The training set consisted of 9,127 images, and the development set consisted of 1,229 images. The network was trained using a binary cross-entropy loss function with a stochastic gradient descent optimizer. The optimizer had a learning rate of 1E-4 and momentum of 0.9. The model was trained until the accuracy on the development set decreased for 8 consecutive epochs, then the weights were reset to those of the epoch with maximum accuracy on the development set. The images were converted to gray-scale then rescaled and randomly cropped down to $224 \times 224$ pixels. Image augmentation through rotations was attempted, but it reduced performance. We hypothesize that this was because subtle perspective rotation can be a differentiator between the driver and front passenger seats. 

\begin{figure}[h]
    \centering
    
    \begin{subfigure}{0.33\textwidth}
    \centering
    \includegraphics[width=.8\linewidth]{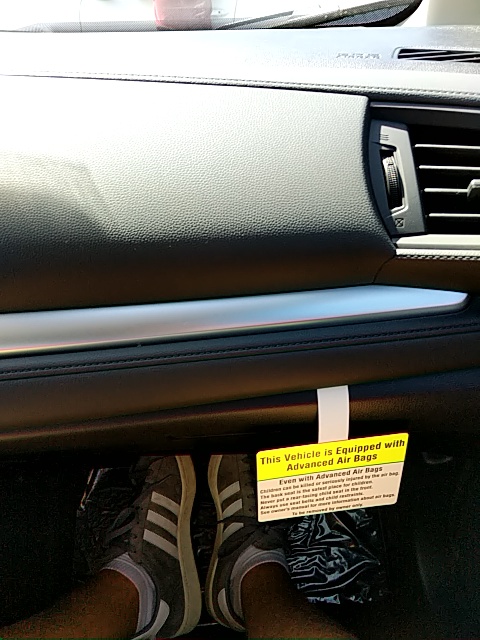}
    \caption{Truth: Passenger \newline Prediction: {\color{green}Passenger}}
    \label{fig:figure1a}
    \end{subfigure}
    \begin{subfigure}{0.33\textwidth}
    \centering
    \includegraphics[width=.8\linewidth]{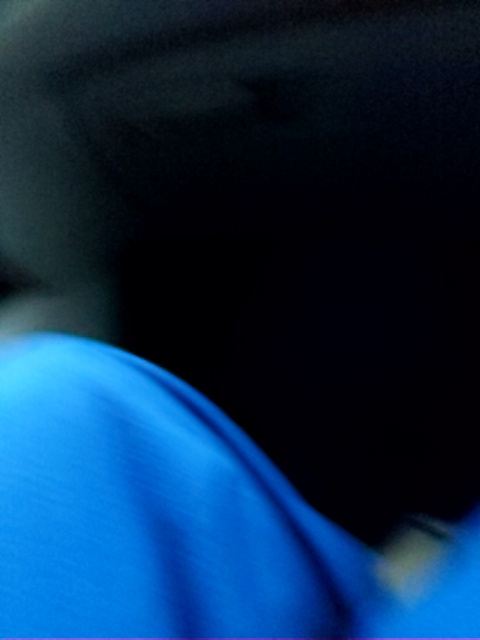}
    \caption{Truth: Passenger \newline Prediction: {\color{green}Passenger}}
    \label{fig:figure1b}
    \end{subfigure}
    \begin{subfigure}{0.33\textwidth}
    \centering
    \includegraphics[width=.8\linewidth]{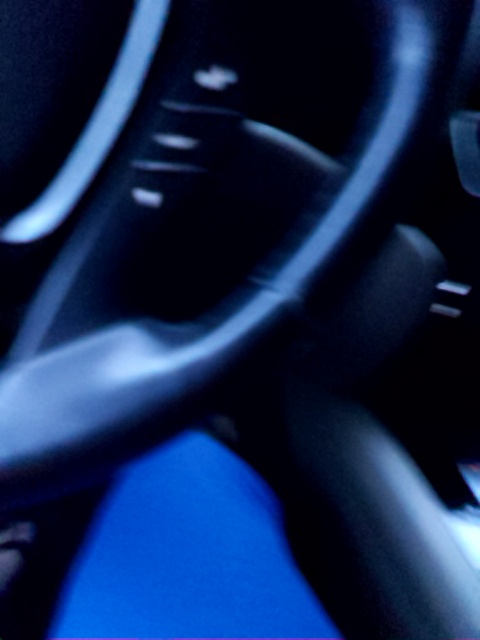}
    \caption{Truth: Driver \newline Prediction: {\color{green}Driver}}
    \label{fig:figure1c}
    \end{subfigure}
    
    \begin{subfigure}{0.33\textwidth}
    \centering
    \includegraphics[width=.8\linewidth]{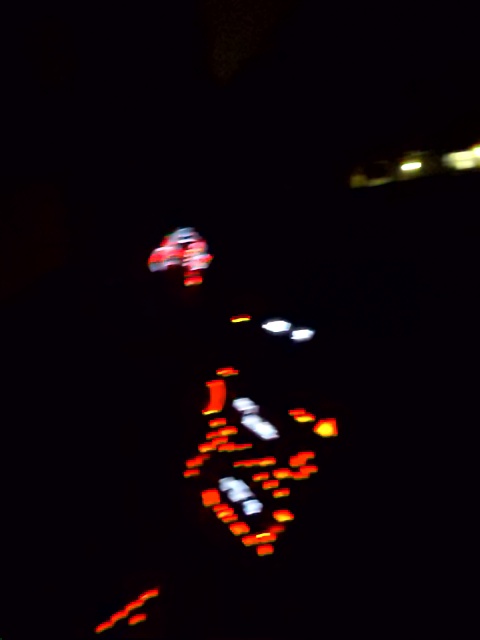}
    \caption{Truth: Passenger \newline Prediction: {\color{red}Driver}}
    \label{fig:figure1d}
    \end{subfigure}
    \begin{subfigure}{0.33\textwidth}
    \centering
    \includegraphics[width=.8\linewidth]{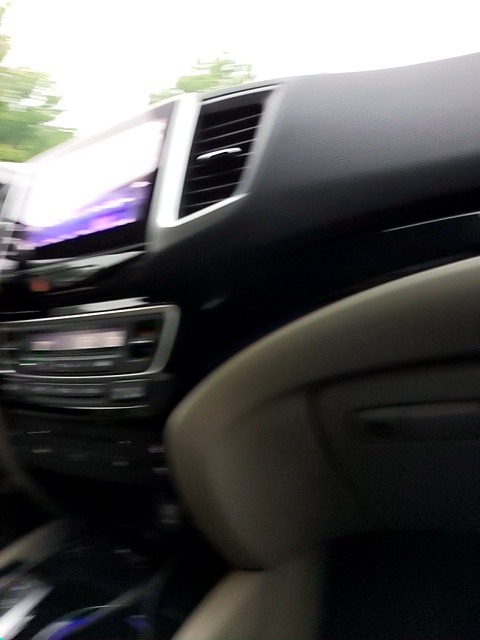}
    \caption{Truth: Passenger \newline Prediction: {\color{red}Driver}}
    \label{fig:figure1e}
    \end{subfigure}
    \begin{subfigure}{0.33\textwidth}
    \centering
    \includegraphics[width=.8\linewidth]{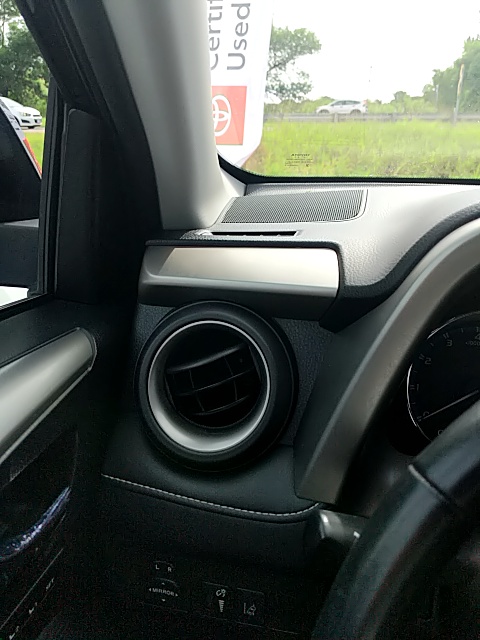}
    \caption{Truth: Driver \newline Prediction: {\color{red}Passenger}}
    \label{fig:figure1f}
    \end{subfigure}
    
    \caption{Example images showing both correct and incorrect predictions in a variety of lighting conditions and image qualities.}
    \label{figure1}
\end{figure}

\section{Results}
The final test set consisted of 1,686 images from a variety of cars, trucks, and lighting conditions. The network was able correctly identify a driver vs. a passenger with an accuracy of 94.9\% when using a 0.5 decision threshold. Examples of correct and incorrect predictions are shown in \textbf{Figure 1}. 

\section{Discussion}
Our results confirm and expand the results of Khurana \textit{et al.} \cite{Khurana2020-qc} and demonstrate the feasibility of using a phone's camera to determine the user's location within a vehicle. Compared to the method in Khurana \textit{et al.}, our method uses only the rear camera, which is typically much better in low light conditions. Our method also uses a purely neural network model compared to a geometric model, so it should generalize better to different vehicles and modes of transportation. This seat detection model could be provided as a library for application developers to use to enable or disable features or to lock the phone completely.

\clearpage

\bibliographystyle{IEEEtran}
\bibliography{references}

\end{document}